%% file: main.tex
\pdfoutput=1
\PassOptionsToPackage{dvipsnames}{xcolor}
\documentclass[11pt]{article}
\usepackage[final]{acl}
\usepackage{times}
\usepackage{latexsym}
\usepackage[T1]{fontenc}
\usepackage[utf8]{inputenc}
\usepackage{microtype}
\usepackage{inconsolata}
\usepackage{graphicx}
\input{math_commands.tex}

\usepackage{url}            

\usepackage{breakurl}
\usepackage[breaklinks]{hyperref}
\usepackage{booktabs}       
\usepackage{amsfonts}       
\usepackage{nicefrac}       
\usepackage{xcolor}         
\usepackage{amsmath,amssymb}
\usepackage{tabulary,multirow,overpic}
\usepackage{wrapfig,lipsum,booktabs}
\usepackage{floatrow}
\usepackage{xspace}
\definecolor{darkgreen}{RGB}{0,112,0}
\definecolor{darkred}{RGB}{112,0,0}
\definecolor{citecolor}{HTML}{0071BC}
\definecolor{linkcolor}{HTML}{ED1C24}
\newcommand{\method}{MobileQuant\xspace}

\title{\method{}: Mobile-friendly Quantization for On-device Language Models}

\author{Fuwen Tan$^1$, Royson Lee$^{1,2}$,  {\L}ukasz Dudziak$^{1,2}$, Shell Xu Hu$^1$, {\bf Sourav Bhattacharya$^1$}, \\ {\bf Timothy Hospedales$^{1,3}$}, {\bf Georgios Tzimiropoulos$^{1,4}$}, {\bf Brais Martinez$^1$} \\ 
Samsung AI Center, Cambridge$^1$ \quad University of Cambridge$^2$ \\ University of Edinburgh$^3$ \quad Queen Mary University of London$^4$ \\ 
\texttt{\{fuwen.tan, royson.lee, l.dudziak, shell.hu, sourav.b1,} \\
\texttt{t.hospedales, georgios.t, brais.mart\}@samsung.com}
}

\begin{document}
\maketitle
\begin{abstract}
Large language models (LLMs) have revolutionized language processing, delivering outstanding results across multiple applications. 
However, deploying LLMs on edge devices poses several challenges with respect to memory, energy, and compute costs, limiting their widespread use in devices such as mobile phones.
A promising solution is to reduce the number of bits used to represent weights and activations.
While existing works have found partial success at quantizing LLMs to lower bitwidths, \textit{e.g.} 4-bit weights, quantizing activations beyond 16 bits often leads to large computational overheads due to poor on-device quantization support, or a considerable accuracy drop. Yet, 8-bit activations are very attractive for on-device deployment
as they would enable LLMs to fully exploit mobile-friendly hardware, e.g. Neural Processing Units (NPUs).
In this work, we make a first attempt to facilitate the on-device deployment of LLMs using integer-only quantization. 
We first investigate the limitations of existing quantization methods for on-device deployment, with a special focus on activation quantization.
We then address these limitations by introducing a simple post-training quantization method, named \method{}, that extends previous weight equivalent transformation works by jointly optimizing the weight transformation and activation range parameters in an end-to-end manner.
\method{} demonstrates superior capabilities over existing methods by 1) achieving near-lossless quantization on a wide range of LLM benchmarks, 2) reducing latency and energy consumption by 20\%-50\% compared to current on-device quantization strategies, 3) requiring limited compute budget, 4) being compatible with mobile-friendly compute units, \textit{e.g.} NPU.
\end{abstract}

\input{01_intro}
\input{02_related_work}
\input{03_method}
\input{04_experiment}

\section{Conclusion}
We revisited LLM quantization from the perspective of deployment on edge devices such as mobile phones. We examined the limitations of current state-of-the-art models for on-device deployment and present \method{}, the first framework to facilitate compute-, and energy-efficient quantized LLMs with minimal performance loss. \method{} is drop-in compatible with today's edge device hardware and low-level runtimes.

\section*{Limitations}
The work explores reducing the overhead of on-device deployment for Large Language Models by hardware-friendly quantization. 
Our current study focuses on established pretrained LLMs with 1 to 2 billion parameters, which limits the overall capacity of the quantized models. Also, the quantized models inherit the error of the pretrained models, e.g. hallucination, which may be corrected by extra guard models~\cite{llamaguard}. For now, we demonstrate the efficiency and effectiveness of \method{} on specific high-end mobile phones.
We plan to extend our research to more LLMs with different architectures, model sizes, capacities, as well as more edge devices in the future.

\bibliography{custom}

\appendix
\clearpage
\input{appendix}

\end{document}

%% file: math_commands.tex
\usepackage{amsmath,amsfonts,bm}

\def\eqref#1{equation~\ref{#1}}

\def\1{\bm{1}}

\DeclareMathAlphabet{\mathsfit}{\encodingdefault}{\sfdefault}{m}{sl}
\SetMathAlphabet{\mathsfit}{bold}{\encodingdefault}{\sfdefault}{bx}{n}



%% file: 01_intro.tex
\section{Introduction}
\label{sec:intro}

Large language models (LLMs) have markedly advanced language processing capabilities, paving the way for expansive applications in artificial intelligence. 
However, the deployment of LLMs is costly in terms of memory, computation, and energy, which can be prohibitive on edge devices like mobile phones. 
A standard approach to facilitate running these models on edge devices is to quantize them, representing weights and activations with fewer bits, thereby mitigating these costs. 

Existing LLM quantization works can be grouped into two categories: \textit{weight-only quantization} and \textit{weight-activation quantization}.
Weight-only quantization approaches~\cite{gptq, awq} convert model weights into low-bitwidth integers, most commonly 4-bit, and maintain the activations in 16-bit floating-point. 
Weight-only quantization often preserves accuracy while significantly reducing the model storage footprint. 
In addition, weight-only quantization can result in minor gains in inference latency due to the reduction in memory access overheads. 
However, these approaches still suffer from high energy consumption and high latency, as computation is performed in floating point. Costly on-the-fly weight dequantization is also required during inference. 
Instead, weight-activation quantization approaches forgo the need for on-the-fly dequantization by quantizing both weights and activations, and potentially utilizing efficient fixed-point operators. 
Despite its efficiency benefits, quantizing activations typically degrades accuracy given the activation outliers~\cite{smoothquant, wu2024fast, luo2024outeffhop}, especially in the case where static per-tensor quantization parameters are applied.
To counteract this accuracy drop, previous works include quantizing activations for certain expensive operations~\cite{smoothquant},\textit{ e.g.} matrix multiplication, or employing dynamic per-token quantization~\cite{omniquant, llmqat, quarot, QLLM}, which is often slow on Graphic Processing Units (GPUs) and, most importantly, lacks hardware support on edge devices. 
Notably, none of these methods support lossless 8-bit (int8) per-tensor quantization for the activations, or fully leverage low-precision fixed-point engines, such as the Digital Signal Processor (DSP), or dedicated Neural Processing Unit (NPU)~\cite{npu, edgetpu}, commonly found in mobile devices~\cite{hexagon}. 
Towards on-device quantization for LLMs, we introduce \method{}, a post-training quantization approach that not only effectively handles the conventional accuracy and efficiency challenges of quantization but is also seamlessly supported by existing mobile hardware.
To achieve this, \method{} consists of three simple yet effective methodological extensions, motivated by the shortcomings of existing state-of-the-art works when deployed on device, and building on top of these works. These extensions include: \textit{1)} applying weight equivalent transformation on \textit{all possible layers}, \textit{2)}, learning the optimal quantization range for activations, \textit{3)} jointly optimizing all weight transformation and range parameters in an end-to-end manner. 
As such, \method{} applies a combination of per-tensor and per-channel weight quantization at 4-bit or 8-bit and per-tensor activation quantization at 8-bit or 16-bit, utilizing fixed-point integer representations for all operations.

The benefits of \method{} over previous works are multifold. Firstly, \method{} enables the quantization of the weights to either 4-bit or 8-bit and the activations to 8-bit integers, except for non-linearities like softmax and normalization, with minimal impact on performance. 
\method{}, hence, maximizes the potential of equivalent transformation-based methods~\cite{equal2019, smoothquant, awq, omniquant} that achieve linear-invariant weight equalization. 
Deploying LLMs on device using \method{} results in a significant reduction in inference speed and energy usage as the latency and energy consumption of multiply-accumulate operations correlate directly with the bit-widths. 
Besides substantial gains during inference, we also show that \method{}'s end-to-end optimization benefits from more calibration samples and extended training samples through our ablation study. 
In contrast, previous works that adopt closed-form solutions~\cite{equal2019}, search-based optimization~\cite{awq}, and block-wise error minimization~\cite{omniquant, QLLM} struggle to scale with the number of samples and training steps. 
Lastly, in comparison with other learnable-based quantization methods such as Quantization Aware Training (QAT)~\cite{llmqat, quantizable}, \method{} retains the model generalizability as the model remains mathematically equivalent to its unquantized variant. Our contributions are summarized as follows:

\begin{enumerate}
\item We introduce a post-training quantization approach for large language models (LLMs) that is supported by current mobile hardware implementations (i.e. DSP, NPU), thus being directly deployable on real edge devices. 
\item Our method improves upon prior works through simple yet effective methodological extensions that enable us to effectively quantize most activations to a lower bitwidth (\textit{i.e.} 8-bit) with near-lossless performance.
\item We conduct a comprehensive on-device evaluation of model accuracy, inference latency, and energy consumption. Our results indicate that our method reduces both inference latency and energy usage by 20\%-50\% while still maintaining accuracy compared to models using 16-bit activations.
\end{enumerate}

%% file: 02_related_work.tex
\section{Related Work}
\label{sec:related_work}

\subsection{Post-training Quantization (PTQ)}

Previous research in post-training quantization for LLMs can be categorized into three main groups:

\noindent\textbf{Weight-only Quantization} focuses on compressing the model weights to reduce storage requirements and memory transfer overheads. 
Representative works~\cite{gptq, awq, omniquant, QLLM} generally achieve performance comparable to full-precision models and maintain similar inference speeds on GPUs. 
However, these methods dequantize weights to 16-bit values on the fly, resulting in high-precision floating-point computations and hence leading to high inference latency and energy consumption, particularly on edge devices such as mobile phones.

\noindent\textbf{Weight-activation Quantization} extends quantization to both model weights and activations,
aiming to further reduce computational overhead. However, as indicated in prior works~\cite{llmint8, smoothquant}, activations have dynamic ranges across different data distributions and are hence more challenging to quantize compared to weights. 
As a result, quantizing activations to a lower bit-width often results in a significant performance decline. 
Leading solutions either retain some compute-intensive matrix multiplications in full precision~\cite{llmint8, smoothquant} or utilize dynamic per-token activation quantization, which lacks hardware support on mobile platforms. 
In contrast, our approach quantizes all linear operations and is compatible with current hardware support on edge devices. 

\noindent\textbf{Learning to Round}. Notable works like~\cite{adaround, flexround} also focus on weight-only quantization but introduce techniques for learning optimal weight rounding. The key argument is that the conventional round-to-nearest method is suboptimal, as it does not account for the interdependencies among adjacent weights. Our work is orthogonal with this research and can hence be integrated with these techniques.

\subsection{Quantization Aware Training (QAT)}

Quantization aware training (QAT) involves retraining or fine-tuning full-precision models using differentiable quantizers. Recent research~\cite{llmqat, quantizable} has shown that QAT outperforms PTQ methods, particularly with in-domain training data. However, QAT requires extensive training, which is often impractical for LLMs. Additionally, QAT may be vulnerable to domain shifts if the data used for pretraining is unavailable. In contrast, our approach is zero-shot, only requiring a minimal set of calibration samples and a limited compute budget. Once trained, our model remains mathematically equivalent to the original model when unquantized, enhancing its adaptability to various downstream tasks.

%% file: 03_method.tex
\section{Preliminaries}\label{sec:preliminaries}

\begin{figure*}[t]
    \centering
    \includegraphics[width=1.0\linewidth]{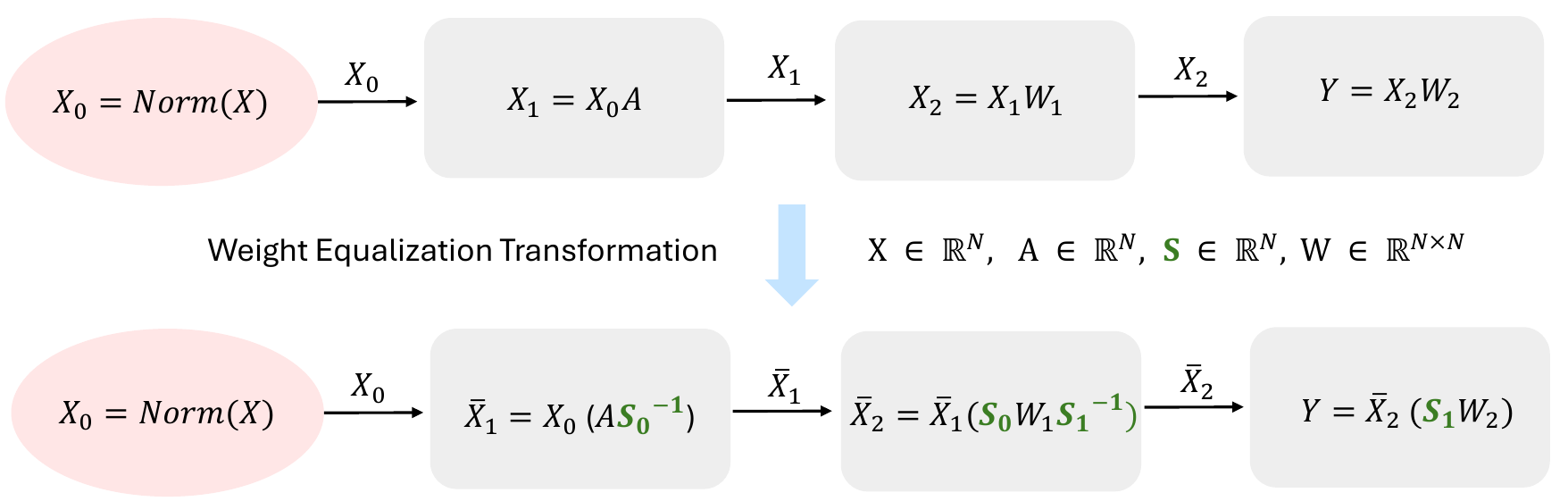}
    \caption{Weight equalization transformation proposed in~\cite{equal2019, smoothquant, omniquant}. In this example, we use three consecutive layers: one normalization layer, \textit{e.g.} LayerNorm~\cite{layernorm}/RMSNorm~\cite{rmsnorm}, and two linear layers, and assume the activations of all layers have the same hidden dimension $N$. Here $A \in \mathbb{R}^N$ refers to the affinity transformation of the normalization layer. The goal of weight transformation is to learn the scaling vector $S$ such that the resulting weight matrices (i.e. $S_0W_1S_1^{-1}$ and $S_1W_2$) and activations $\bar{X}$, are easier to quantize. $S$ is hence the only learnable parameters. Note that the new model is mathematically equivalent to the original model when unquantized.} 
    \label{fig:trans}
\end{figure*}

\subsection{Mobile-friendly Design Choices}

Quantization methods are differentiated by several main design choices, with varying levels of hardware support. In this section, we first list these design choices and then highlight the limitations of existing works with respect to these choices. 

\noindent\textbf{Support for mobile-friendly bitwidth:} int8-int8 operations are widely supported and most often optimized for, while int4-int16 and int8-int16 are typically supported although often slower than int8-int8. 

\noindent\textbf{Quantization groups:} Quantizing using per-tensor and per-channel statistics is widely supported while using per-token statistics is not.

\noindent\textbf{Dynamic vs static:} Static quantization statistics that do not depend on the input data, typically computed on a holdout calibration set, are widely supported. Dynamic quantization, on the other hand, requires online calibration from the input data and is not widely supported.

State-of-the-art quantization methods demonstrate strong performance on a server use case (i.e. high-end GPU). However, they either utilize on-the-fly dequantization and 16-bit floating point operations~\cite{gptq, awq}, which are computationally inefficient, or dynamic per-token quantization~\cite{smoothquant, omniquant}, which, as previously mentioned, has no support on edge devices.

We, instead, consider design choices that are widely supported and optimized on modern edge devices (\textit{e.g.} Mobile NPUs), namely \textit{i)} fixed-point weight and activation quantization with integer arithmetic operations, and \textit{ii)} per-tensor/channel quantization with static pre-computed ranges. Our objective is hence to improve existing state-of-the-art approaches such as SmoothQuant~\cite{smoothquant} and OmniQuant~\cite{omniquant} while staying within the limits of hardware support on device.

\subsection{Weight Equivalent Transformation}\label{sec:weight_equivalent_transformation}

Prior efforts on LLM quantization~\cite{llmint8, smoothquant} observed that activations are harder to quantize compared to the model weights due to the outlier channel dimensions with diverse min-max ranges. As an example, given a fully connected layer $\mathbf{Y} = \mathbf{XW}$, $\mathbf{W} \in \mathbb{R}^{N \times M}, \mathbf{X} \in \mathbb{R}^{N}, \mathbf{Y} \in \mathbb{R}^{M}$, specific channel dimensions $\{i: 0 \leq i < N\}$ in $\mathbf{X}$ may have a wide min-max range across different data samples, causing substantial quantization errors. To counteract this, previous methods proposed a weight equivalent transform defined by a scaling vector $\mathbf{S} \in \mathbb{R}^{N}$:

\begin{eqnarray}
\mathbf{Y} = \mathbf{XW} = (\mathbf{X}\mathbf{S}^{-1})\cdot (\mathbf{SW}) = \hat{\mathbf{X}}\hat{\mathbf{W}}
\end{eqnarray}

The goal is to find the optimal scaling vector $\mathbf{S}$ such that both $\hat{\mathbf{X}}$ and $\hat{\mathbf{W}}$ are easier to quantize compared to the original $\mathbf{X}$ and $\mathbf{W}$. SmoothQuant~\cite{smoothquant} reparameterized $\mathbf{S}$ as $\mathbf{s}_i = \frac{max(|\mathbf{X}_i|)^\alpha}{max(|\mathbf{W}_i|)^{(1-\alpha)}})$, $0 \leq i < N$, and searched for the hyper-parameter $\alpha$. The obtained $\mathbf{S}$ is similar to the closed-form solution derived in~\cite{equal2019}. OmniQuant~\cite{omniquant} extended SmoothQuant~\cite{smoothquant} by learning $\mathbf{S}$, together with the weight clipping parameters via block-wise error minimization. Here, both $\mathbf{S}$ and $\mathbf{S}^{-1}$ can be fused to the adjacent linear layers, making the transformation mathematically equivalent to the original models. Figure~\ref{fig:trans} provides an illustration of the transformation among consecutive linear layers.




\section{\method{}: Towards Mobile-friendly Quantization}
\label{sec:method}

\subsection{Challenges for Mobile-friendly Quantization}\label{sec:our_weight_equivalent_transformation} 
The weight equivalent transformation approaches used in SmoothQuant and OmniQuant, as described in Section~\ref{sec:weight_equivalent_transformation},
demonstrate strong performance on GPU-like hardware. However, they do not work out of the box for edge devices. 
Specifically, two challenges remain: 
i) the weight transformations cannot propagate beyond non-linear operators, \textit{e.g.} Softmax, RMSNorm~\cite{rmsnorm}, LayerNorm~\cite{layernorm}, SiLU/GELU~\cite{gelu}. To counteract this, we apply weight transformations on all consecutive layers with linear components, \textit{e.g.} between linear layers or affine transformations in the normalization layers, while keeping the nonlinear activations in 16-bit integers; 
ii) with the weight transformation, the distribution of the activations shifts accordingly. This causes essential difficulty for learning-based approaches like OmniQuant~\cite{omniquant}, when the min-max range for the activations changes after each training iteration. OmniQuant~\cite{omniquant} proposed to bypass the issue with dynamic per-token quantization, which has no hardware support on-device.

\subsection{Learning the Per-tensor Range of the Activations}
\label{method:range}
Given the distribution of the activations shifts accordingly with the weight transformation, the ideal solution is to re-estimate the activation ranges across the training set after each training iteration. However, doing so is computationally prohibited. Hence, we propose to learn the activation range jointly with the weight transformation.
Given an activation tensor $\mathbf{X}$, instead of learning the min and max values $f_{min}(\mathbf{X})$, $f_{max}(\mathbf{X})$ directly, we leverage the correlation between $f_{min}$, $f_{max}$ and the scale and offset parameters, $\alpha, \beta \in \mathbb{R}$, for quantization. With the targeted bit-width $bw$, quantizing $\mathbf{X}$ can be formulated as: 

\begin{align}
q_{max} = &2^{bw} - 1, \\
\textcolor{ForestGreen}{\alpha} = \frac{\textcolor{darkred}{f_{max}}-\textcolor{darkred}{f_{min}}}{q_{max}}&,
\textcolor{ForestGreen}{\beta} = \frac{\textcolor{darkred}{f_{min}}}{\textcolor{ForestGreen}{\alpha}} \\
\mathbf{X}_{int} = min(max(ste&(\frac{\mathbf{X}}{\textcolor{ForestGreen}{\alpha}}) - \textcolor{ForestGreen}{\beta}, 0), q_{max})
\end{align}

Here, $\mathbf{X}_{int}$ refers to the quantized tensor of $\mathbf{X}$, $ste$ refers to straight-through estimator. We can therefore learn $\textcolor{darkred}{f_{min}} = \textcolor{ForestGreen}{\alpha} \textcolor{ForestGreen}{\beta}$ and $\textcolor{darkred}{f_{max}} = \textcolor{ForestGreen}{\alpha} q_{max} + \textcolor{ForestGreen}{\alpha} \textcolor{ForestGreen}{\beta}$ indirectly by learning $\textcolor{ForestGreen}{\alpha}$ and $\textcolor{ForestGreen}{\beta}$, which are computationally more stable. 


\subsection{End-to-end Optimization vs Layer-wise Optimization}

To learn the equivalent transformation, previous works either resort to closed-form solutions~\cite{equal2019}, search-based methods~\cite{smoothquant, awq}, or layer-wise error minimization~\cite{omniquant}. These solutions require limited training budget, but, as shown in Section.~\ref{subsec:e2e}, lead to sub-optimal performance. Particularly, given the restricted form of supervision, we show that these methods cannot scale with more training samples or iterations. We, instead, propose to jointly optimize all the training parameters, including the weight equalization parameters $\mathbf{S}$, weight clipping parameters used in OmniQuant~\cite{omniquant}, and the range parameters $\alpha, \beta$ for all layers in an end-to-end manner. Compared to previous PTQ approaches, which struggle with more training samples and epochs, we demonstrate that our holistic optimization approach consistently improves the performance with larger training settings for different LLM architectures. Compared to QAT, our method preserves model generalizability and does not overfit to specific calibration samples, achieving near-lossless zero-shot performance.

%% file: 04_experiment.tex
\section{Experiments}
\label{sec:exp}

\begin{table}[t]
\centering
\scalebox{0.76}{
    \input{tables/baseline}
}
    \caption{\textbf{Adapting quantization SOTA to the on-device setting}. OmniQuant and SmoothQuant are not fully supported for on-device deployment. We introduce mobile-friendly variants. Evaluation: perplexity on WikiText~\cite{wikitext}.
    We adopt the ``Edge'' variants as strong on-device baselines.}
    \label{tab:baseline} 
\end{table}

\subsection{Setup}
\label{subsec:setting}

We perform experiments by training and simulating the quantization on GPUs and further evaluate the on-device performance on a Samsung Galaxy S24, with the Snapdragon 8 Gen 3 HTP as the compute unit. All models were trained on two A100 GPUs, with a maximum sequence length of 2048.

\noindent\textbf{Architectures:} \method{} focuses on lightweight LLMs that are suitable to be deployed on mobile devices. Hence, we experiment with representative pretrained models with different architectures: 
TinyLlaMA-1.1B-Chat-v1.0~\cite{tinyllama}, 
StableLM-2-1.6B~\cite{stablelm}, 
and Gemma-2B~\cite{gemma}.\\

\noindent\textbf{Quantization details.}
\method{} use a subset of the Pile~\cite{pile} dataset as the calibration set. We explore two quantization settings: i) W8A8: 8-bit weight quantization with per-tensor statistics except for the last linear projection in each MLP block (e.g. $down\_proj$ in LLaMA-like~\cite{llama} models) which uses per-channel statistics, and 8-bit per-tensor quantization for the activations, except those linked to non-linear operators. ii) W4A8: 4-bit per-channel quantization for model weights, and 8-bit per-tensor quantization, likewise excluding non-linear operators. \\
We consider asymmetric quantization for both settings, which can utilize the full quantized range. 
We also provide extra experiments on symmetric per-channel W4A8 quantization in the supplemental material, which is better supported by the current on-device toolchain we use.

\noindent\textbf{Evaluation datasets.}
We evaluate our quantization approach in a zero-shot setting on representative tasks from the Language Model Evaluation Harness benchmark (Harness)~\cite{eval-harness} including WikiText~\cite{wikitext}, AI2 Reasoning Challenge (arc\_challenge)~\cite{arc_challenge}, Hellaswag~\cite{hellaswag}, and MMLU~\cite{mmlu}.

\begin{table*}[t]
\centering
\scalebox{0.95}{
    \input{tables/e2e}
}
\caption{\textbf{End-to-end range optimization:} Perplexity on WikiText for OmniQuant-Edge W4A8 setting with block-wise vs end-to-end range optimization. Best overall performance is in \textbf{bold}, best block-wise performance is \underline{underlined}. Compared to block-wise, end-to-end optimization benefits from larger training settings with more samples/iterations, leading to better performance.}
\label{tab:e2e}
\end{table*}

\subsection{On-device Baselines}
\label{subsec:baseline}

In this section, we extend state-of-the-art weight-activation quantization methods, SmoothQuant~\cite{smoothquant} and OmniQuant~\cite{omniquant} on device and use them as baselines.
As these approaches utilize dynamic per-token quantization for the activation, which is not supported on edge devices, we modify these methods to work on device by using static per-tensor activation quantization, referring to these variants as OmniQuant-Static and SmoothQuant-Static respectively. 
Note that, for SmoothQuant, we only include evaluations on W8A8, which is the default setting used in the original work.

As shown in Table~\ref{tab:baseline}, both ``Static'' variants suffer from large performance degradation when evaluated on WikiText~\cite{wikitext}.
We further observe that the performance drop is mainly caused by quantizing the activations for the last linear layer in each MLP head (\textit{i.e}. $\mathit{down\_proj}$ in LLaMA-like~\cite{llama} models). To further alleviate this issue, we introduce an extra weight equalization transformation between consecutive linear layers in each MLP head (\textit{i.e}. $\mathbf{S}$ between the $\mathit{up\_proj}$ and $\mathit{down\_proj}$ layers in TinyLLaMA~\cite{tinyllama}).
The new models, which we termed \textit{SmoothQuant-Edge} and \textit{OmniQuant-Edge} respectively, significantly alleviate the performance degradation. 
For the remainder of this section, we use these adapted models as strong on-device baselines.

\begin{table}[h]
    \scalebox{0.71}{
        \input{tables/range}
    }
    \caption{\textbf{Activation range learning} (ARL): Perplexity on WikiText for OmniQuant-Edge with/without ARL for W8A8 and W4A8 settings. The performance gains are larger on models with larger quantization errors.}
    \label{tab:range}
\end{table}

\subsection{Impact of Activation Range Learning}
\label{subsec:range}
Table~\ref{tab:baseline} shows that the learning-based approach, OmniQuant~\cite{omniquant}, outperforms the search-based method, SmoothQuant~\cite{smoothquant}, for all models by a notable margin. 
However, learning to transform the weights with fixed activation ranges is suboptimal, as the activation ranges shift after each training iteration.
We further evaluate the impact of incorporating activation range learning (ARL), described in Section.~\ref{method:range}, into OmniQuant~\cite{omniquant}. 
In other words, we learn the per-tensor scale and offset parameters, together with the weight transformation via block-wise error minimization.

Table~\ref{tab:range} demonstrates that activation range learning (ARL) consistently improves the performance for all LLM models across all settings. The gains are larger for quantized models exhibiting a larger performance gap compared to the FP16 models. 
Notably, these models require more training steps to mitigate the quantization errors, leading to larger range shifts for the activation.

\subsection{Impact of End-to-End Optimization}
\label{subsec:e2e}

In the previous section, we show that incorporating ARL into our baselines results in consistent improvements. Nonetheless, there is still a notable performance gap between the quantized models and the FP16 models, especially under the W4A8 setting. In order to reduce this gap, we attempt to improve the performance by scaling up the performance, namely increasing the number of calibration samples and the number of training epochs. However, Table~\ref{tab:e2e} shows that the performance of all considered models saturate as we scale the training up using the block-wise approach proposed in OmniQuant~\cite{omniquant}.
We therefore conjecture that the optimization is hindered by the block-wise error minimization objective that provides limited global supervision.
To verify this, we use our end-to-end training pipeline and jointly optimize all trainable parameters of the whole model, namely the weight transformation, clipping, and activation range learning parameters.

\begin{table*}[t]
    \centering
\scalebox{0.96}{
    \input{tables/harness}
}
    \caption{\textbf{Comparisons with existing state-of-the-art methods on Harness:} 
    Best performance is \textbf{bold}, second-best underlined. We indicate the \textcolor{ForestGreen}{gain}/\textcolor{darkred}{drop} of our approach vs the \underline{next strongest on-device baseline}. Our method, \method{}, demonstrates consistent improvements across models, quantization configurations, and tasks, achieving best performance in most cases.}
    \label{tab:harness}
\end{table*}

As shown in Table~\ref{tab:e2e}, our end-to-end trained models demonstrate consistent improvements with more training samples and iterations, only underperforming the blockwise optimized models in the smallest settings when the models were undertrained.
We currently train the models with up to 1024 samples for 60 epochs but posit that the models could be further improved with more diverse samples and larger training settings.

\subsection{Harness Benchmark Results}
\label{subsec:harness}

Following previous approaches~\cite{smoothquant, omniquant, llmqat}, we perform zero-shot evaluations on representative tasks from the Harness benchmark~\cite{eval-harness}.
%
Table~\ref{tab:harness} shows that, in addition to the WikiText perplexity, our method also improves the quantization performance for the common sense reasoning tasks in general, without using any in-domain data. The improvements are consistent for most benchmarks and we believe that the performance of our method could be further improved with in-domain data, especially for benchmarks with a large domain shift relative to our calibration set (i.e. Pile~\cite{pile}).

\begin{table}[t]
    \centering
    \scalebox{0.95}{
        \input{tables/accuracy}
    }
    \caption{\textbf{On-device accuracy} of the quantized TinyLLaMA-1.1B-Chat-v1.0 on WikiText and LAMBADA. Models run on a Snapdragon 8 Gen 3 HTP processor. }
    \label{tab:ondevice_accuracy}
\end{table}

\begin{table*}[t]
    \centering
    \scalebox{1.0}{

\input{tables/energy}
    }
    \caption{\textbf{On-device execution cost.} Measurements of latency, energy and memory are computed under sustained execution (30 minutes). Values are reported per single forward pass.}
    \label{tab:energy_and_memory}
\end{table*}

\subsection{On-device Evaluation}
\label{subsec:ondevice}

\noindent \textbf{On-device Setup. } 
We further deploy the quantized LLM model on a mobile device and provide evaluations on the accuracy, latency, memory usage, and power consumption.
Specifically, we evaluate the W8A8 quantized TinyLLaMA-1.1B-Chat-v1.0~\cite{tinyllama} model on a Samsung Galaxy S24, using the Snapdragon 8 Gen 3 HTP as the compute unit.
We evaluate the model under three different quantization settings: 1) \textit{W8A16}, which keeps activations as 16-bit; Note that the matrix multiplication for the self-attention computation is still between 8-bit and 16-bit unsigned integer activations to avoid potential overflowing, 2) \textit{full W8A8}, keeps all activations in 8-bit, and 4) our proposed \method{} for W8A8. 

\noindent \textbf{On-device Accuracy. } 
We compute the accuracy of the quantized models on two tasks: i) WikiText~\cite{wikitext} from Harness~\cite{eval-harness}, as we used in our previous evaluations and ii) LAMBADA~\cite{lambada}, which predicts the last token of a sentence given the previous context. Following SmoothQuant~\cite{smoothquant}, we use the first 1000 samples from LAMBADA for this task. Table~\ref{tab:ondevice_accuracy} shows that using 16-bit activations (i.e. \textit{W8A16}) achieves lossless performance. However, quantizing all activations into 8-bit leads to near-zero performance, highlighting the difficulty of activation quantization. Our W8A8 \method{} model achieves near-lossless performance in both tasks, approaching the performance of the FP16 model.

\noindent \textbf{On-device Latency. }
We provide the on-device latency evaluation by running the quantized model in two modes: i) prompt encoding with a context length of 256, ii) auto-regressive generation with a maximum sequence length of 1024 and 2048. Table~\ref{tab:energy_and_memory} shows that, for prompt encoding, using lower-bitwidth activations is critical to reducing the inference latency, as some of the operations, e.g. self-attention (batched matrix multiplication), are compute-intensive. Our model demonstrates significant advantages over the full W8A16 solution, reducing the latency by 40\%. However, there is still a large gap between \method{} and the \textit{full W8A8} model, indicating the improvement margin. For auto-regressive generation, the latency gaps are smaller. 
We posit that the auto-regressive generation is not as compute-bound as prompt encoding, especially for lightweight models, but instead is partially memory access-bound. Our solution demonstrates a 20\% latency reduction compared to \textit{W8A16}, achieving the same latency as the \textit{full W8A8} model. 
We include a video demo that showcases the auto-regressive generation of the quantized model on device in the supplemental material.
In general, the advantage of low-bitwidth activations correlates strongly with the scale of the computation.
Hence, we aim to extend the latency evaluation to larger models in our future research. 


\noindent \textbf{On-device Energy and Memory.}
Apart from latency, energy consumption is another important aspect of on-device execution, which is often overlooked by quantization research.
To measure the energy requirements of different models, we run them on a number of identical mobile phones as used before continuously for 30 minutes.
The phones are connected to the testing host machine via WiFi using an internal network without access to the internet, to avoid any undesired network activity.
The phones are also not being charged and their screens are turned off.
All phones begin each test at the same battery level and the final energy of running a model is calculated as the ratio of the total battery discharged over the duration of a test, minus reference discharge of a phone not running any model, divided by the number of times the model was run.
We repeat measurements for different models 3 times, rotating the phones each time, and report the average.
We also report peak memory required to run a model as the peak resident memory recorded for the benchmarking process by the Linux Kernel (the so-called Virtual Memory High Water Mark).
From Table~\ref{tab:energy_and_memory}, the energy consumption of each model aligns well with the latency. 
Compared to \textit{W8A16}, \method{} reduces 50\% of the power usage for prompt encoding and 35\% for autoregressive generation. The peak memory usage for all models are similar as it is dominated by the model weight. 

%% file: tables/baseline.tex

\begin{tabular}{l|l|l|l}
    \toprule
    \multirow{2}{*}{ WikiText ($\downarrow$) } & TinyLLaMA & StableLM-2 & Gemma \\
    & 1.1B & 1.6B & 2B \\
    \midrule
    {FP16} & 14.9 & 28.4 & 18.0 \\
    \midrule
    \multicolumn{4}{c} {\textbf{W8A8}}\\
    \midrule
    SmoothQuant-Static & 177 & 583 & >1E+03 \\
    SmoothQuant-Edge & 27.1 & 74.5	& 45.3 \\
    OmniQuant-Static & 51.0	& 298.6 & >1E+03 \\
    OmniQuant-Edge & 16.3 & 30.9 & 23.4 \\
    \midrule
    \multicolumn{4}{c} {\textbf{W4A8}}\\
    \midrule
    OmniQuant-Static & 416.3 & 258.5 & >1E+03 \\
    OmniQuant-Edge & 18.8 & 36.0 & 23.9 \\
    \bottomrule
\end{tabular}

%% file: tables/e2e.tex
\begin{tabular}{ll|ll|ll|ll}
    \toprule
    \multirow{2}{*} {\#Samples} & \multirow{2}{*} {\#Epochs}  & \multicolumn{2}{c|} {TinyLlaMA-1.1B} & \multicolumn{2}{c|} {StableLM-2-1.6B}  & \multicolumn{2}{c} {Gemma-2B}  \\
     &   & Block-wise & End-to-end  &  Block-wise & End-to-end &  Block-wise & End-to-end \\
    \midrule
      128   & 20 & 18.3 & 19.9 & \underline{35.4} & 40.4 & \underline{23.0} & 32.5  \\
      128   & 60 & 18.3 & 17.4 & 37.0 & 36.5 & 23.7 & 26.1 \\
      128   & 120 & 18.1 & \textbf{17.1} & 37.1 & 35.1  & 24.0 &  23.1 \\
      256   & 60 & 17.9 & \textbf{17.1} & 35.9 & 34.2 & 24.5 & 22.0\\
      1024  & 60 &  \underline{17.7} & \textbf{17.1} & \underline{35.4} &  \textbf{33.6} & 24.9 & \textbf{21.4} \\
    \bottomrule
\end{tabular}

%% file: tables/range.tex

\begin{tabular}{l|l|l|l}
    \toprule
    \multirow{2}{*}{ WikiText ($\downarrow$) } & TinyLLaMA & StableLM-2 & Gemma \\
    & 1.1B & 1.6B & 2B \\
    \midrule
    {FP16} & 14.9 & 28.4 & 18.0 \\
    \midrule
    \multicolumn{4}{c} {\textbf{W8A8}}\\
    \midrule 
    OmniQuant-Edge & 16.3 & 30.9 & 23.4 \\
    OmniQuant-Edge w ARL & 15.9 & 30.5 & 22.8 \\
    \midrule
    \multicolumn{4}{c} {\textbf{W4A8}}\\
    \midrule
    OmniQuant-Edge & 18.8 & 36.0 & 23.9 \\
    OmniQuant-Edge w ARL & 18.3 & 35.4 & 23.0 \\
    \bottomrule
\end{tabular}

%% file: tables/harness.tex
\begin{tabular}{l|l|l|l|l|l}
    \toprule
     \multicolumn{2}{l|} {} & WikiText $\downarrow$ & ARC-Challenge $\uparrow$ & HellaSwag $\uparrow$ & MMLU $\uparrow$ \\ 
     \midrule
     \multicolumn{6}{c} {\textbf{W8A8}}  \\
     \midrule
    \multirow{4}{*}{TinyLlaMA-1.1B} & FP16  & 14.9 & 33 & 60  & 25 \\ 
        & SmoothQuant-Edge  & 27.1 & 29.6 & 52.8 & 24.9 \\
     & OmniQuant-Edge  & \underline{16.3}  & \underline{31.7} & \underline{58.4} & 24.9\\
     & \method{} & \textbf{15.5} (\footnotesize{\textcolor{ForestGreen}{-0.8}}) & \textbf{31.9} (\footnotesize{\textcolor{ForestGreen}{+0.2}})& \textbf{59.2} (\footnotesize{\textcolor{ForestGreen}{+0.8}})  & \textbf{25.0} (\footnotesize{\textcolor{ForestGreen}{+0.1}})\\
     \midrule
     \multirow{4}{*}{StableLM-2-1.6B} & FP16  & 28.4  & 39  & 65 & 32 \\ 
        & SmoothQuant-Edge   & 70.2 & 35.9  & 61.8 &26.0\\
     & OmniQuant-Edge & \underline{30.9} & \underline{36.3} & \underline{63.4} & \underline{29.3}\\
     & \method{} & \textbf{29.7} (\footnotesize{\textcolor{ForestGreen}{-1.2}}) & \textbf{37.1} (\footnotesize{\textcolor{ForestGreen}{+0.8}}) & \textbf{63.6} (\footnotesize{\textcolor{ForestGreen}{+0.2}})  & \textbf{30.0} (\footnotesize{\textcolor{ForestGreen}{+0.7}})\\
     \midrule
     \multirow{4}{*}{Gemma-2B} & FP16  & 18.0  & 23  & 42  & 28 \\ 
        & SmoothQuant-Edge &45.3 & \textbf{23.0} & 39.0 & 25.8\\
     & OmniQuant-Edge   & \underline{23.4} & \underline{22.4}  & \underline{39.9} & \underline{26.8}\\
     & \method{} & \textbf{20.3} (\footnotesize{\textcolor{ForestGreen}{-3.1}}) & 21.8 (\footnotesize{\textcolor{darkred}{-1.2}})  & \textbf{40.9} (\footnotesize{\textcolor{ForestGreen}{+1.0}})  & \textbf{25.8} (\footnotesize{\textcolor{darkred}{-1.0}}) \\
     \midrule
     \multicolumn{6}{c} {\textbf{W4A8}}  \\
     \midrule
    \multirow{3}{*}{TinyLlaMA-1.1B} & FP16  & 14.9 & 33 & 60  & 25 \\ 
     & OmniQuant-Edge   & \underline{18.8} & \underline{28.8} & \underline{56.4} & \textbf{25.5}\\
     & \method{} & \textbf{17.1} (\footnotesize{\textcolor{ForestGreen}{-1.7}}) & \textbf{32.3} (\footnotesize{\textcolor{ForestGreen}{+3.5}})  & \textbf{57.0} (\footnotesize{\textcolor{ForestGreen}{+0.6}})   & \textbf{25.5} (\footnotesize{\textcolor{ForestGreen}{+0.0}}) \\
     \midrule
     \multirow{3}{*}{StableLM-2-1.6B} & FP16  & 28.4  & 39  & 65 & 32 \\ 
     & OmniQuant-Edge & \underline{36.0} & 34.9 & 60.2  & 25.9 \\
     & \method{} & \textbf{33.6} (\footnotesize{\textcolor{ForestGreen}{-2.4}}) & \textbf{35.6} (\footnotesize{\textcolor{ForestGreen}{+0.7}}) & \textbf{60.5} (\footnotesize{\textcolor{ForestGreen}{+0.3}}) & \textbf{24.1} (\footnotesize{\textcolor{darkred}{-1.8}})\\
     \midrule
     \multirow{3}{*}{Gemma-2B} & FP16  & 18.0  & 23  & 42  & 28 \\ 
     & OmniQuant-Edge & \underline{23.9}  & \textbf{23.1} & \underline{38.1} & \underline{25.5} \\
     & \method{} & \textbf{21.4} (\footnotesize{\textcolor{ForestGreen}{-2.5}}) & \underline{23.0} (\footnotesize{\textcolor{darkred}{-0.1}}) & \textbf{38.9} (\footnotesize{\textcolor{ForestGreen}{+0.8}}) & \textbf{25.6} (\footnotesize{\textcolor{ForestGreen}{+0.1}}) \\
     \bottomrule
\end{tabular}

%% file: tables/accuracy.tex
\begin{tabular}{l|l|l}
\toprule
TinyLlaMA-1.1B & WikiText $\downarrow$ & Lambada $\uparrow$ \\
\midrule
FP16 & 14.9 & 82.9 \\
\textit{W8A16} & 15.2 & 82.9 \\ 
\method{} \textit{W8A8} & 15.6 & 82.4 \\
\textit{full W8A8} & 8e5 & 1.3 \\
\bottomrule
\end{tabular}

%% file: tables/energy.tex
\begin{tabular}{l|l|c|c|c}
    \toprule
    Seq. Length    & Method   & Avg. lat. (ms)   & Avg. energy (mJ)  & Peak mem. (MiB) \\ \midrule
    \multicolumn{5}{c}{\textbf{Prompt Encoding}} \\\midrule
    \multirow{3}{*}{256}    & \textit{W8A16}             & 510   & 1000   & 1019 \\
                            & \method{} (\textit{W8A8})         & 276   & 490   & 1011 \\ 
                            & \textit{full W8A8}              & 89   & 183   & 1006 \\
    \midrule
    \multicolumn{5}{c}{\textbf{Autoregressive Generation}} \\ \midrule
    \multirow{3}{*}{1024}   & \textit{W8A16}             & 54   & 69   & 1007 \\
                            & \method{} (\textit{W8A8})        & 46   & 61   & 1005 \\ 
                            & \textit{full W8A8}             & 42   & 61   & 1003 \\
    \midrule
    \multirow{3}{*}{2048}   & \textit{W8A16}             & 119   & 165   & 1010 \\
                            & \method{} (\textit{W8A8})         & 95   & 110   & 1007 \\
                            & \textit{full W8A8}             & 94   & 106   & 1006 \\
    \bottomrule
\end{tabular}

%% file: appendix.tex
\section{Appendix: On-device Experiments for W4A8}
\label{sec:appendix}

In this appendix, we provide further on-device evaluation for W4A8.
The current on-device toolchains we use support only symmetric per-channel weight quantization.
This, however, typically leads to performance degradation as the full quantization range may not be fully utilized if the weights are biased toward positive or negative.
Here we first present extra W4A8 results with symmetric per-channel quantization.
We then include on-device latency evaluation showcasing the advantages of using 4-bit integer representation for the weights.

\subsection{Symmetric vs Asymmetric W4A8 Quantization}
We train extra W4A8 models with symmetric per-channel quantization.
Table~\ref{tab:symm} presents the performance of symmetric per-channel W4A8 models on Wikitext~\cite{eval-harness}, confirming the performance degradation compared to the asymmetric counterparts. 

\begin{table}[h]
    \centering
    \scalebox{0.75}{
        \input{tables/symm}
    }
    \caption{Evaluation of symmetric vs asymmetric W4A8 per-channel quantization on Wikitext~\cite{eval-harness}.}
    \label{tab:symm}
\end{table}

\begin{table}[h]
    \centering
    \scalebox{0.8}{
        \input{tables/latency}
    }
    \caption{On-device latency (ms) for TinyLlaMA-1.1B~\cite{tinyllama} and Gemma-2B~\cite{gemma} across different settings.}
    \label{tab:latency}
\end{table}

\subsection{On-device Latency for Symmetric W4A8 models}

We further evaluate the on-device latency of the W4A8 models with symmetric quantization. Table\ref{tab:latency} shows that, compared to W8A8, the W4A8 models demonstrate improved inference speed for both prompt encoding and autoregressive generation. For larger models like Gemma-2B, the improvements are more significant, \textit{i.e.} reducing the latency of prompt encoding and autoregressive generation by 39\% and 33\%. Here, TinyLLaMA-1.1B achieves the same inference speed, \textit{i.e.} 40 ms per token (25 tok/s). We conjecture that, in this setting, the autoregressive generation for these models is likely memory-bound. We plan to further investigate the performance bottleneck in future research.

%% file: tables/symm.tex
\begin{tabular}{l|l|l|l}
    \toprule
    \multirow{2}{*}{ WikiText ($\downarrow$) } & TinyLLaMA & StableLM-2 & Gemma \\
    & 1.1B & 1.6B & 2B \\
    \midrule
    {FP16} & 14.9 & 28.4 & 18.0 \\
    \midrule
    MobileQuant-Asym & 17.1 & 33.6 & 21.4 \\
    MobileQuant-Sym & 17.5 & 36.4 & 24.7 \\
    \bottomrule
\end{tabular}

%% file: tables/latency.tex


\begin{tabular}{l|l|l}
    \toprule
     Method & TinyLlaMA-1.1B & Gemma-2B \\
     \midrule
     \multicolumn{3}{c}{Prompt Encoding (Seq. Length 256)} \\
     \midrule
     \textit{W8A16} & 510 & 1191\\
     \method{} (\textit{W8A8}) & 276 & 752 \\
     \textit{full W8A8} & 89 & 311 \\
     \midrule
     \textit{W4A16} & 320 & 617\\
     \method{} (\textit{W4A8}) & 239 & 460 \\
     \textit{full W4A8} & 89 & 98 \\
     \midrule
     \multicolumn{3}{c}{Autoregressive Gen. (Context Length 1024)} \\
     \midrule
     \textit{W8A16} & 54 & 78\\
     \method{} (\textit{W8A8}) & 46 & 60 \\
     \textit{full W8A8} & 42 & 59 \\
     \midrule
     \textit{W4A16} & 50 & 56\\
     \method{} (\textit{W4A8})& 38 & 40 \\
     \textit{full W4A8} & 40 &  40 \\
    \bottomrule
\end{tabular}


%% file: main.bbl
\begin{thebibliography}{32}
\providecommand{\natexlab}[1]{#1}

\bibitem[{Ashkboos et~al.(2024)Ashkboos, Mohtashami, Croci, Li, Jaggi, Alistarh, Hoefler, and Hensman}]{quarot}
Saleh Ashkboos, Amirkeivan Mohtashami, Maximilian~L. Croci, Bo~Li, Martin Jaggi, Dan Alistarh, Torsten Hoefler, and James Hensman. 2024.
\newblock \href {https://arxiv.org/abs/2404.00456} {{QuaRot}: Outlier-free 4-bit inference in rotated llms}.
\newblock \emph{Preprint}, arXiv:2404.00456.

\bibitem[{Ba et~al.(2016)Ba, Kiros, and Hinton}]{layernorm}
Jimmy~Lei Ba, Jamie~Ryan Kiros, and Geoffrey~E. Hinton. 2016.
\newblock \href {https://arxiv.org/abs/1607.06450} {Layer normalization}.
\newblock \emph{Preprint}, arXiv:1607.06450.

\bibitem[{Bellagente et~al.(2024)Bellagente, Tow, Mahan, Phung, Zhuravinskyi, Adithyan, Baicoianu, Brooks, Cooper, Datta et~al.}]{stablelm}
Marco Bellagente, Jonathan Tow, Dakota Mahan, Duy Phung, Maksym Zhuravinskyi, Reshinth Adithyan, James Baicoianu, Ben Brooks, Nathan Cooper, Ashish Datta, et~al. 2024.
\newblock \href {https://arxiv.org/abs/2402.17834} {{Stable LM} 2 1.6 b technical report}.
\newblock \emph{Preprint}, arXiv:2402.17834.

\bibitem[{Bondarenko et~al.(2023)Bondarenko, Nagel, and Blankevoort}]{quantizable}
Yelysei Bondarenko, Markus Nagel, and Tijmen Blankevoort. 2023.
\newblock Quantizable transformers: Removing outliers by helping attention heads do nothing.
\newblock In \emph{Advances on Neural Information Processing Systems}.

\bibitem[{Clark et~al.(2018)Clark, Cowhey, Etzioni, Khot, Sabharwal, Schoenick, and Tafjord}]{arc_challenge}
Peter Clark, Isaac Cowhey, Oren Etzioni, Tushar Khot, Ashish Sabharwal, Carissa Schoenick, and Oyvind Tafjord. 2018.
\newblock \href {https://arxiv.org/abs/1803.05457} {Think you have solved question answering? try {ARC}, the {AI2} reasoning challenge}.
\newblock \emph{Preprint}, arXiv:1803.05457.

\bibitem[{Dettmers et~al.(2022)Dettmers, Lewis, Belkada, and Zettlemoyer}]{llmint8}
Tim Dettmers, Mike Lewis, Younes Belkada, and Luke Zettlemoyer. 2022.
\newblock {LLM.int8()}: 8-bit matrix multiplication for transformers at scale.
\newblock In \emph{Advances on Neural Information Processing Systems}.

\bibitem[{Frantar et~al.(2023)Frantar, Ashkboos, Hoefler, and Alistarh}]{gptq}
Elias Frantar, Saleh Ashkboos, Torsten Hoefler, and Dan Alistarh. 2023.
\newblock {GPTQ}: Accurate post-training compression for generative pretrained transformers.
\newblock In \emph{International Conference on Learning Representations}.

\bibitem[{Gao et~al.(2020)Gao, Biderman, Black, Golding, Hoppe, Foster, Phang, He, Thite, Nabeshima, Presser, and Leahy}]{pile}
Leo Gao, Stella Biderman, Sid Black, Laurence Golding, Travis Hoppe, Charles Foster, Jason Phang, Horace He, Anish Thite, Noa Nabeshima, Shawn Presser, and Connor Leahy. 2020.
\newblock \href {https://arxiv.org/abs/2101.00027} {The {P}ile: An {800GB} dataset of diverse text for language modeling}.
\newblock \emph{Preprint}, arXiv:2101.00027.

\bibitem[{Gao et~al.(2023)Gao, Tow, Abbasi, Biderman, Black, DiPofi, Foster, Golding, Hsu, Le~Noac'h, Li, McDonell, Muennighoff, Ociepa, Phang, Reynolds, Schoelkopf, Skowron, Sutawika, Tang, Thite, Wang, Wang, and Zou}]{eval-harness}
Leo Gao, Jonathan Tow, Baber Abbasi, Stella Biderman, Sid Black, Anthony DiPofi, Charles Foster, Laurence Golding, Jeffrey Hsu, Alain Le~Noac'h, Haonan Li, Kyle McDonell, Niklas Muennighoff, Chris Ociepa, Jason Phang, Laria Reynolds, Hailey Schoelkopf, Aviya Skowron, Lintang Sutawika, Eric Tang, Anish Thite, Ben Wang, Kevin Wang, and Andy Zou. 2023.
\newblock \href {https://doi.org/10.5281/zenodo.10256836} {A framework for few-shot language model evaluation}.

\bibitem[{Google(2021)}]{edgetpu}
Google. 2021.
\newblock Edge tpu.
\newblock \url{https://cloud.google.com/edge-tpu}.

\bibitem[{Google(2024)}]{gemma}
Google. 2024.
\newblock \href {https://arxiv.org/abs/2403.08295} {Gemma: Open models based on {Gemini} research and technology}.
\newblock \emph{Preprint}, arXiv:2403.08295.

\bibitem[{Hendrycks et~al.(2021)Hendrycks, Burns, Basart, Zou, Mazeika, Song, and Steinhardt}]{mmlu}
Dan Hendrycks, Collin Burns, Steven Basart, Andy Zou, Mantas Mazeika, Dawn Song, and Jacob Steinhardt. 2021.
\newblock Measuring massive multitask language understanding.
\newblock \emph{International Conference on Learning Representations}.

\bibitem[{Hendrycks and Gimpel(2016)}]{gelu}
Dan Hendrycks and Kevin Gimpel. 2016.
\newblock \href {https://arxiv.org/abs/1606.08415} {Bridging nonlinearities and stochastic regularizers with gaussian error linear units}.
\newblock \emph{Preprint}, arXiv:1606.08415.

\bibitem[{Inan et~al.(2023)Inan, Upasani, Chi, Rungta, Iyer, Mao, Tontchev, Hu, Fuller, Testuggine, and Khabsa}]{llamaguard}
Hakan Inan, Kartikeya Upasani, Jianfeng Chi, Rashi Rungta, Krithika Iyer, Yuning Mao, Michael Tontchev, Qing Hu, Brian Fuller, Davide Testuggine, and Madian Khabsa. 2023.
\newblock \href {https://arxiv.org/abs/2312.06674} {Llama guard: {LLM}-based input-output safeguard for human-{AI} conversations}.
\newblock \emph{Preprint}, arXiv:2312.06674.

\bibitem[{Lee et~al.(2023)Lee, Kim, Kwon, and Lee}]{flexround}
Jung~Hyun Lee, Jeonghoon Kim, Se~Jung Kwon, and Dongsoo Lee. 2023.
\newblock {FlexRound}: Learnable rounding based on element-wise division for post-training quantization.
\newblock In \emph{International Conference on Machine Learning}.

\bibitem[{Lin et~al.(2024)Lin, Tang, Tang, Yang, Chen, Wang, Xiao, Dang, Gan, and Han}]{awq}
Ji~Lin, Jiaming Tang, Haotian Tang, Shang Yang, Wei-Ming Chen, Wei-Chen Wang, Guangxuan Xiao, Xingyu Dang, Chuang Gan, and Song Han. 2024.
\newblock {AWQ}: Activation-aware weight quantization for llm compression and acceleration.
\newblock In \emph{Conference on Machine Learning and Systems}.

\bibitem[{Liu et~al.(2024)Liu, Gong, Wei, Dong, Cai, and Zhuang}]{QLLM}
Jing Liu, Ruihao Gong, Xiuying Wei, Zhiwei Dong, Jianfei Cai, and Bohan Zhuang. 2024.
\newblock Qllm: Accurate and efficient low-bitwidth quantization for large language models.
\newblock In \emph{The Twelfth International Conference on Learning Representations}.

\bibitem[{Liu et~al.(2023)Liu, Oguz, Zhao, Chang, Stock, Mehdad, Shi, Krishnamoorthi, and Chandra}]{llmqat}
Zechun Liu, Barlas Oguz, Changsheng Zhao, Ernie Chang, Pierre Stock, Yashar Mehdad, Yangyang Shi, Raghuraman Krishnamoorthi, and Vikas Chandra. 2023.
\newblock {LLM-QAT}: Data-free quantization aware training for large language models.
\newblock \emph{arXiv preprint arXiv:2307.06281}.

\bibitem[{Luo et~al.(2024)Luo, Hu, Chang, Chen, Li, Wang, and Liu}]{luo2024outeffhop}
Haozheng Luo, Jerry Yao-Chieh Hu, Pei-Hsuan Chang, Hong-Yu Chen, Weijian Li, Wei-Po Wang, and Han Liu. 2024.
\newblock \href {https://openreview.net/forum?id=ZCrRCICOkr} {Outeffhop: A principled outlier-efficient attention layer from dense associative memory models}.
\newblock In \emph{Workshop on Efficient Systems for Foundation Models II @ ICML2024}.

\bibitem[{Mahurin(2023)}]{hexagon}
E.~Mahurin. 2023.
\newblock \href {https://doi.org/10.1109/HCS59251.2023.10254715} {Qualocmm® hexagon™ {NPU}}.
\newblock In \emph{IEEE Hot Chips Symposium}, pages 1--19. IEEE Computer Society.

\bibitem[{Merity et~al.(2016)Merity, Xiong, Bradbury, and Socher}]{wikitext}
Stephen Merity, Caiming Xiong, James Bradbury, and Richard Socher. 2016.
\newblock \href {https://arxiv.org/abs/1609.07843} {Pointer sentinel mixture models}.
\newblock \emph{CoRR}, abs/1609.07843.

\bibitem[{Nagel et~al.(2020)Nagel, Amjad, Van~Baalen, Louizos, and Blankevoort}]{adaround}
Markus Nagel, Rana~Ali Amjad, Mart Van~Baalen, Christos Louizos, and Tijmen Blankevoort. 2020.
\newblock Up or down? {A}daptive rounding for post-training quantization.
\newblock In \emph{International Conference on Machine Learning}.

\bibitem[{Nagel et~al.(2019)Nagel, van Baalen, Blankevoort, and Welling}]{equal2019}
Markus Nagel, Mart van Baalen, Tijmen Blankevoort, and Max Welling. 2019.
\newblock Data-free quantization through weight equalization and bias correction.
\newblock In \emph{IEEE International Conference on Computer Vision}.

\bibitem[{Paperno et~al.(2016)Paperno, Kruszewski, Lazaridou, Pham, Bernardi, Pezzelle, Baroni, Boleda, and Fern{\'a}ndez}]{lambada}
Denis Paperno, Germ{\'a}n Kruszewski, Angeliki Lazaridou, Ngoc~Quan Pham, Raffaella Bernardi, Sandro Pezzelle, Marco Baroni, Gemma Boleda, and Raquel Fern{\'a}ndez. 2016.
\newblock The {LAMBADA} dataset: Word prediction requiring a broad discourse context.
\newblock In \emph{Annual Meeting of the Association for Computational Linguistics}.

\bibitem[{Qualcomm(2024)}]{npu}
Qualcomm. 2024.
\newblock Unlocking on-device generative ai with an npu and heterogeneous computing.
\newblock \url{https://www.qualcomm.com/content/dam/qcomm-martech/dm-assets/documents/Unlocking-on-device-generative-AI-with-an-NPU-and-heterogeneous-computing.pdf}.

\bibitem[{Shao et~al.(2024)Shao, Chen, Zhang, Xu, Zhao, Li, Zhang, Gao, Qiao, and Luo}]{omniquant}
Wenqi Shao, Mengzhao Chen, Zhaoyang Zhang, Peng Xu, Lirui Zhao, Zhiqian Li, Kaipeng Zhang, Peng Gao, Yu~Qiao, and Ping Luo. 2024.
\newblock {OmniQuant}: Omnidirectionally calibrated quantization for large language models.
\newblock In \emph{International Conference on Learning Representations}.

\bibitem[{Touvron et~al.(2023)Touvron, Lavril, Izacard, Martinet, Lachaux, Lacroix, Rozière, Goyal, Hambro, Azhar, Rodriguez, Joulin, Grave, and Lample}]{llama}
Hugo Touvron, Thibaut Lavril, Gautier Izacard, Xavier Martinet, Marie-Anne Lachaux, Timothée Lacroix, Baptiste Rozière, Naman Goyal, Eric Hambro, Faisal Azhar, Aurelien Rodriguez, Armand Joulin, Edouard Grave, and Guillaume Lample. 2023.
\newblock \href {https://arxiv.org/abs/2302.13971} {{LLaMA}: Open and efficient foundation language models}.
\newblock \emph{Preprint}, arXiv:2302.13971.

\bibitem[{Wu et~al.(2024)Wu, Lu, Luo, Hu, Wang, Liu, Dehak, Villalba, and Liu}]{wu2024fast}
Shang Wu, Yen-Ju Lu, Haozheng Luo, Jerry Yao-Chieh Hu, Jiayi Wang, Jing Liu, Najim Dehak, Jesus Villalba, and Han Liu. 2024.
\newblock \href {https://openreview.net/forum?id=kZBM2UYiQh} {Fast adaptation and robust quantization of multi-modal foundation models from associative memory: A case study in speech{LM}}.
\newblock In \emph{Workshop on Efficient Systems for Foundation Models II @ ICML2024}.

\bibitem[{Xiao et~al.(2023)Xiao, Lin, Seznec, Wu, Demouth, and Han}]{smoothquant}
Guangxuan Xiao, Ji~Lin, Mickael Seznec, Hao Wu, Julien Demouth, and Song Han. 2023.
\newblock {S}mooth{Q}uant: Accurate and efficient post-training quantization for large language models.
\newblock In \emph{International Conference on Machine Learning}.

\bibitem[{Zellers et~al.(2019)Zellers, Holtzman, Bisk, Farhadi, and Choi}]{hellaswag}
Rowan Zellers, Ari Holtzman, Yonatan Bisk, Ali Farhadi, and Yejin Choi. 2019.
\newblock {H}ella{S}wag: Can a machine really finish your sentence?
\newblock In \emph{Annual Meeting of the Association for Computational Linguistics}.

\bibitem[{Zhang and Sennrich(2019)}]{rmsnorm}
Biao Zhang and Rico Sennrich. 2019.
\newblock Root mean square layer normalization.
\newblock In \emph{Advances on Neural Information Processing Systems}.

\bibitem[{Zhang et~al.(2024)Zhang, Zeng, Wang, and Lu}]{tinyllama}
Peiyuan Zhang, Guangtao Zeng, Tianduo Wang, and Wei Lu. 2024.
\newblock \href {https://arxiv.org/abs/2401.02385} {{TinyLlama}: An open-source small language model}.
\newblock \emph{Preprint}, arXiv:2401.02385.

\end{thebibliography}
